\documentclass{article}

\usepackage{PRIMEarxiv}

\usepackage[utf8]{inputenc} 
\usepackage[T1]{fontenc}    
\usepackage{url}            
\usepackage{booktabs}       
\usepackage{amsfonts}       
\usepackage{nicefrac}       
\usepackage{microtype}      
\usepackage{lipsum}
\usepackage{fancyhdr}       
\usepackage{graphicx}       
\graphicspath{{media/}}     
\usepackage{subfigure}
\usepackage{graphicx}
\usepackage{amsmath}
\usepackage{amssymb}
\usepackage{booktabs}
\usepackage{multirow}
\usepackage{algorithm}
\usepackage{algpseudocode}
\usepackage{amsmath}
\usepackage[mathscr]{euscript}
\usepackage{graphbox}
\usepackage{subcaption}

\usepackage[pagebackref,breaklinks,colorlinks,bookmarks=false]{hyperref}

\title{SIDE: Self-supervised Intermediate Domain Exploration for Source-free Domain Adaptation

}

\author{
  Jiamei Liu, Han Sun*,  Yizhen Jia, \\
  Nanjing University of Aeronautics and Astronautics \\
  Nanjing\\
  \texttt{\{jiamei, sunhan, yz.jia\}@nuaa.edu.cn} \\
   \And
  Jie Qin, \\
  Nanjing University of Aeronautics and Astronautics \\
  Nanjing\\
  \texttt{qinjiebuaa@gmail.com} \\
  \AND
  Huiyu Zhou, \\
  University of Leicester \\
  Englend \\
  \texttt{hz143@leicester.ac.uk} \\
  \And
  Ningzhong Liu, \\
  Nanjing University of Aeronautics and Astronautics \\
  Nanjing \\
  \texttt{lnz@163.com} \\
}

\begin{document}
\maketitle

\begin{abstract}
Domain adaptation aims to alleviate the domain shift when transferring the knowledge learned from the source domain to the target domain. Due to privacy issues, source-free domain adaptation (SFDA), where source data is unavailable during adaptation, has recently become very demanding yet challenging. Existing SFDA methods focus on either self-supervised learning of target samples or reconstruction of virtual source data. The former overlooks the transferable knowledge in the source model, whilst the latter introduces even more uncertainty. To address the above issues, this paper proposes self-supervised intermediate domain exploration (SIDE) that effectively bridges the domain gap with an intermediate domain, where samples are cyclically filtered out in a self-supervised fashion. First, we propose cycle intermediate domain filtering (CIDF) to cyclically select intermediate samples with similar distributions over source and target domains. Second, with the aid of those intermediate samples, an inter-domain gap transition (IDGT) module is developed to mitigate possible distribution mismatches between the source and target data. Finally, we introduce cross-view consistency learning (CVCL) to maintain the intrinsic class discriminability whilst adapting the model to the target domain. Extensive experiments on three popular benchmarks, i.e. Office-31, Office-Home and VisDA-C, show that our proposed SIDE achieves competitive performance against state-of-the-art methods. \href{https://github.com/se111/SIDE}{code:https://github.com/se111/SIDE}
\end{abstract}


\section{Introduction}

\begin{figure}[htbp]
\centering
\begin{minipage}[t]{1.0\textwidth}
\centering
\includegraphics[width=0.75\textwidth]{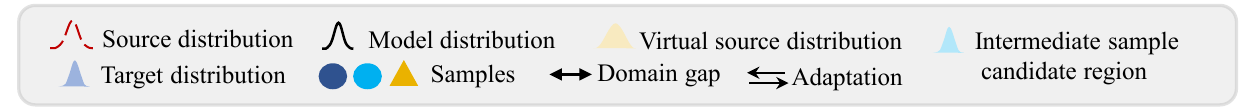}
\end{minipage}

\begin{minipage}[t]{0.75\textwidth}
\centering
\includegraphics[width=0.75\textwidth]{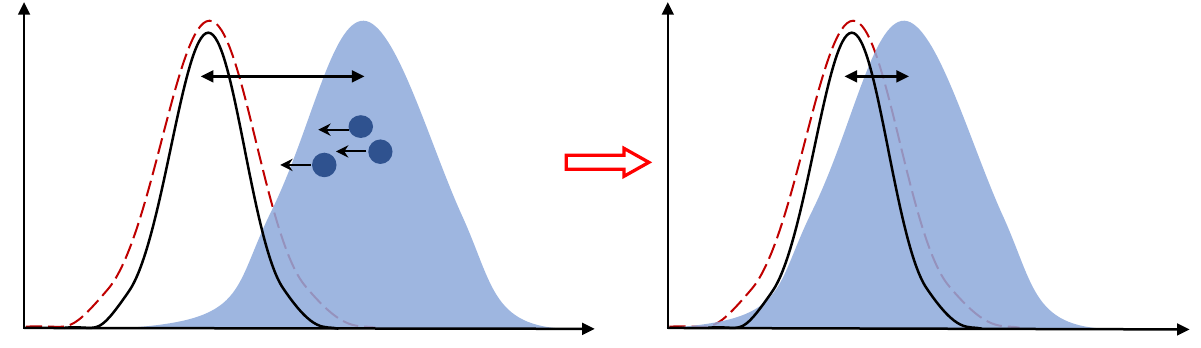}
\text{(a) Self-supervised learning based SFDA methods}
\end{minipage}

\begin{minipage}[t]{0.75\textwidth}
\centering
\includegraphics[width=0.75\textwidth]{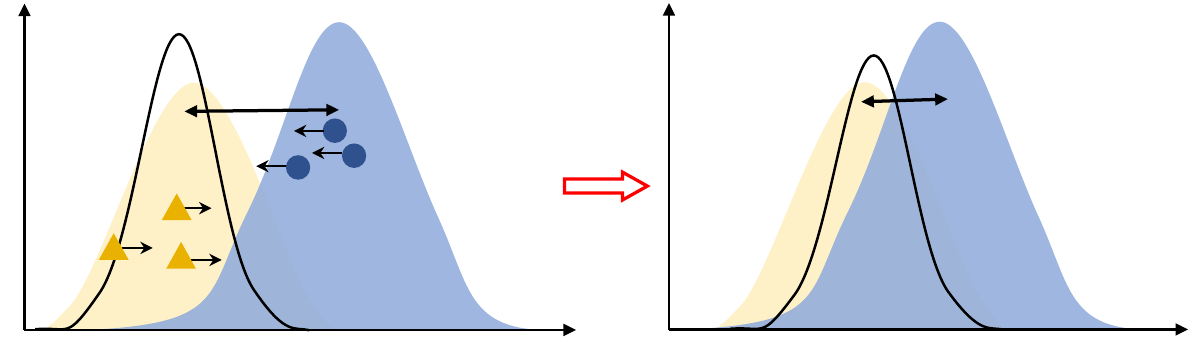}
\text{(b) Virtual source based SFDA methods}
\end{minipage}

\begin{minipage}[t]{0.75\textwidth}
\centering
\includegraphics[width=0.75\textwidth]{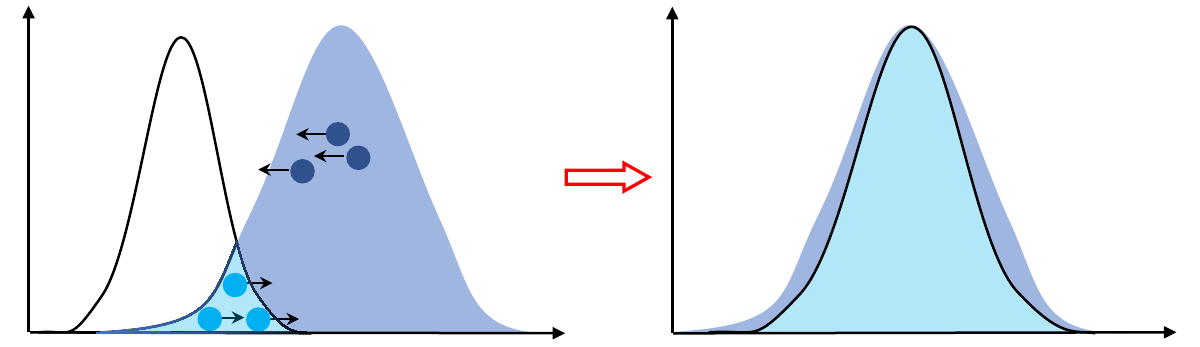}
\text{(c) Our proposed SIDE}
\end{minipage}
\caption{Illustration of different SFDA methods. (a) SSL based methods usually generate pseudo-labels for target data, ignoring the large distribution shift even if the source model well fits the source domain. (b) Virtual source based methods synthesize source data to assist domain adaptation, introducing additional costs and uncertainty. (c) SIDE selects intermediate data from the target domain, gradually adapting the model to the target domain.}
\label{fig1}
\end{figure}

In spite of remarkable successes in various applications, deep neural networks rely on training on a large amount of labeled data. Besides, when applying the trained model to a novel domain, the performance will drop dramatically due to the disagreement between the distributions of the original (source) domain and the new (target) domain, also known as domain shift. To overcome this, unsupervised domain adaptation (UDA) has received a plethora of research interests, aiming to transfer the knowledge learned from the labeled source domain to the unlabeled target domain by alleviating the domain shift. However, the source domain data, which is indispensable for the adaptation process, may not be accessible due to data privacy or intellectual property concerns. To circumvent this limitation, source-free domain adaptation (SFDA) has recently emerged as a more practical setting, where only the pre-trained source model is available without any access to the source data.

To re-adapt the model to the target domain under this challenging setting, a number of SFDA methods have been proposed, which can be mainly divided into the following two categories. The first line of approaches perform self-supervised learning directly based on target samples, as shown in Fig. \ref{fig1} (a). The basic idea is to explore reliable pseudo-labels of target data to calibrate the model \cite{chen2022self,yang2021generalized}. Nevertheless, the quality of pseudo-labels is difficult to guarantee, as the source model is pre-trained based on source data and thus has a limited adaptive ability on the target domain due to the large domain gap. The second line of methods transfer the knowledge from the virtual source domain, where virtual source data is generated by means of synthesis \cite{kurmi2021domain}\cite{qiu2021source} or style transfer \cite{eastwood2021source}\cite{li2021divergence}, as illustrated in Fig. \ref{fig1} (b). However, reconstructing the source domain has two drawbacks. On the one hand, the replacement of the virtual source data has severe uncertainty and there is still a domain gap even if the source data can be perfectly restored. On the other hand, the generative networks involved in source data reconstruction unnecessarily introduce additional computational overhead.

Therefore, to effectively and efficiently tackle SFDA, we need to not only dig out the intrinsic characteristic of the target domain, but also explicitly reveal and eliminate the domain differences without adding new data. To fulfill this goal, in this paper, we propose self-supervised intermediate domain exploration (SIDE) to calibrate the model through self-supervised consistency training with the help of cyclically explored intermediate domain samples. Specifically, we incorporate three carefully-designed modules into SIDE, including cycle intermediate domain filtering (CIDF), inter-domain gap transition (IDGT), and cross-view consistency learning (CVCL). To progressively reduce the gap between source and target domains, CIDF cyclically discovers those target samples similar to source domain prototypes to form the intermediate domain in a self-supervised learning manner. The intuition behind CIDF is that the selected samples work as the bridge to make the distribution of the source domain gradually approach that of the target domain. Furthermore, we propose IDGT to generate augmented samples based on target and intermediate domain samples, properly reflecting the influence of the intermediate domain. Finally, CVCL introduces instance- and class-level consistency losses to guarantee the consistency between two different views of the same target sample.

It is noteworthy that our proposed SIDE shows its unique advantages over the existing SFDA methods. 1) Compared to self-supervised methods, SIDE learns target data representations by leveraging an intermediate domain that shares similarities with both source and target domains. This avoids the interference on the target distribution brought by the inaccurately assigned pseudo labels. Besides, our proposed consistency losses further guarantee the class discriminability, leading to improved classification accuracy. 2) In comparison with `virtual' source data based methods, SIDE does not require any additional time-consuming generation procedures. It leverages `real' intermediate data cyclically to facilitate the transition from the source domain to the target one. In other words, we gradually increase the adaptability of the learned target model, which is often overlooked in previous works.

We summarize our main contributions as follows:

$\bullet$ We tackle the challenging SFDA task by a novel framework, namely self-supervised intermediate domain exploration (SIDE), which exploits intermediate samples cyclically to reduce the cross-domain gap and assist self-supervised learning in the target domain.

$\bullet$ We propose three delicately-designed modules (\emph{i.e.}, CIDF, IDGT, and CVCL), which collaboratively facilitate a smooth adaptation from the pre-trained source model to the target one. As a result, the domain gap is gradually eliminated whilst learning more discriminative target features.


$\bullet$ Extensive experiments on three benchmarks demonstrate that the proposed approach achieves state-of-the-art performance compared to both UDA and SFDA methods.

\section{Related Work}
\subsection{Unsupervised Domain Adaptation}

Early UDA methods mitigate domain shift mainly through instance re-weighting \cite{jiang2007instance} \cite{wang2017instance}\cite{yao2010boosting}
and feature adaptation \cite{gopalan2011domain} \cite{gong2012geodesic}\cite{sun2015subspace}. The former re-weights the source data so that the source distribution can be closer to the target distribution in a non-parametric manner, whereas the gap between two domains is not too significant. Feature adaptation aims to discover the common feature representations of data using metrics \cite{herath2017learning}, projections \cite{si2009bregman} and subspace alignment \cite{liu2019optimal}, \emph{etc}. Benefiting from the development of deep learning, alignment-based methods \cite{ge2017borrowing} \cite{zhang2015deep} \cite{long2015learning}  and adversarial-based methods \cite{zhang2020advkin} \cite{du2021generation} \cite{ganin2015unsupervised} have become popular. However, due to data privacy, the classical UDA method can not adapt to the target domain well with the absence of source domain data.

\begin{figure*}[!t]
\centering
    \includegraphics[width=1.0\textwidth]{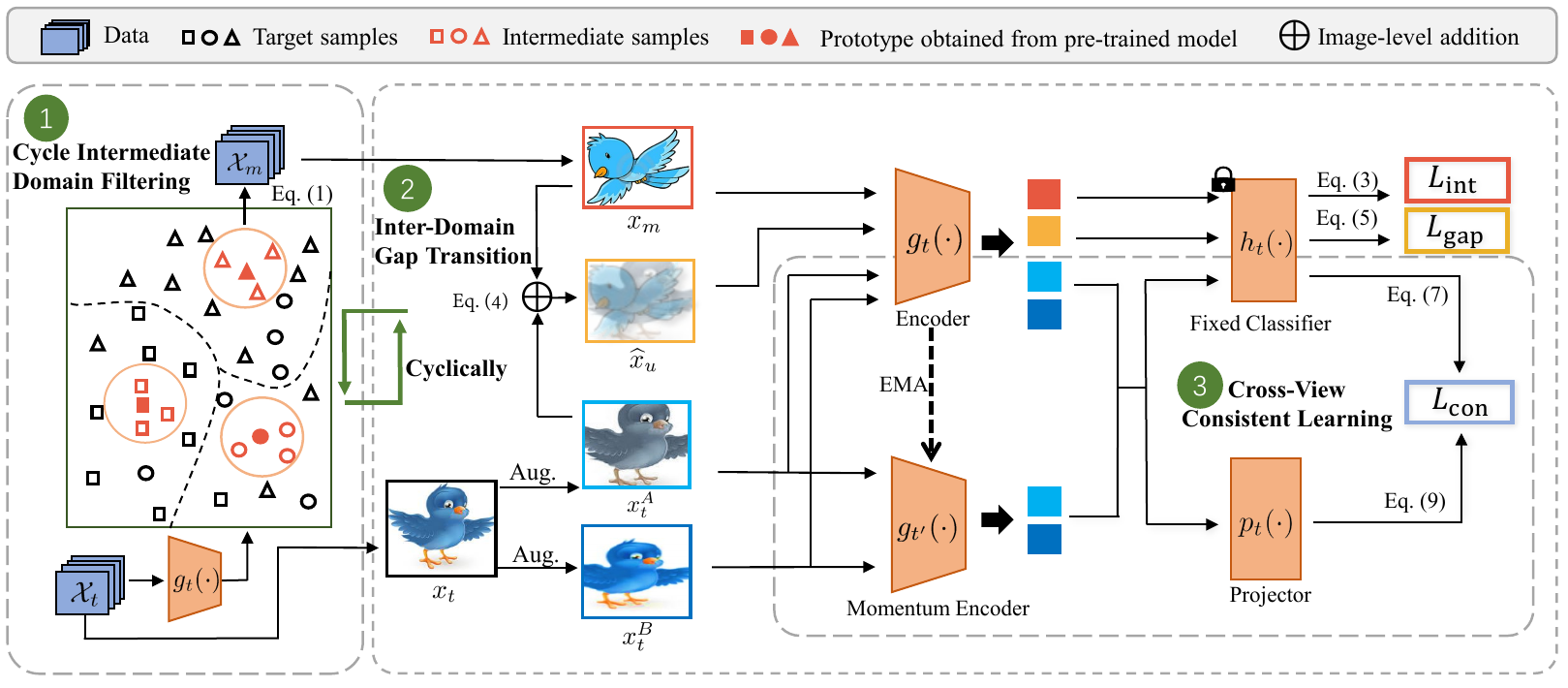}
    \vspace{-2mm}
\caption{The framework of the proposed SIDE. We first select the intermediate data from the target domain by calculating the feature-level similarities between the target samples and the prototypes, and repeat this process throughout the training procedure cyclically. Then, the intermediate and the target samples are trained with the mixup regularization to eliminate the implicit gap. Finally, we optimize the feature encoder by calculating instance- and class-level consistency losses considering different views of a sample. Best viewed in color.}
\label{fig2}
\end{figure*}

\subsection{Source-Free Domain Adaptation}

SFDA aims to adapt the source model to the unlabeled target domain data without accessing the source data. The existing methods are mainly divided into two categories. One is using the methods of pseudo-labels based on self-supervised learning. Liang et al. \cite{liang2021source} maximize mutual information and generate pseudo labels to match the target features to the source hypothesis. Kim et al. \cite{kim2021domain} select samples by entropy values for target model training. However, the accumulation of noise caused by false pseudo-labels brings negative transfer. Different from them, we use the CVCL module to explore deep intrinsic features and corresponding category information.

Some works have attempted to recover the pseudo-source domain by generative methods. Qiu et al. \cite{qiu2021source} generate pseudo-source domain data by additionally training a generator. Du et al. \cite{du2021generation} augment the typical target domain samples. Ding et al. \cite{ding2022proxymix} select the pseudo-source samples based on the feature similarity with the class prototypes, whilst ignoring that the distribution of features learned by the target model is gradually shifted. In this paper, we propose to cyclically choose intermediate domain samples that enables us to push the learnt feature distribution towards the unknown target domain distribution.

\subsection{Self-Supervised Learning in SFDA}
Since there is no human-annotated label in the target domain, self-supervised learning (SSL) has been widely applied in SFDA. Contrastive learning is a common paradigm in SSL with the main idea of minimizing the distance between anchors and negative samples \cite{chen2020improved}\cite{chen2021exploring}. Qiu et al. \cite{qiu2021source} explore the
boundary information of the source model and align the source/target data via contrastive learning. However, pushing away the same-class samples as negative pairs makes the training unstable. Inspired by the approaches without using negative samples \cite{grill2020bootstrap}
\cite{chen2021exploring}, we perform instance- and class-level consistency regularization on cross-views to learn domain-invariant feature representations.

\section{Proposed Method}
\label{sec:formatting}
In this section, we first describe the problem setup of the source-free domain adaptation task. Then, we elaborate our proposed method, namely self-supervised intermediate domain exploration (SIDE) in detail, which is composed of three modules, \emph{i.e.}, cycle intermediate domain filter-
ing (CIDF), inter-domain gap transition (IDGT), and cross-view consistency learning (CVCL). Specifically, CIDF obtains intermediate domain samples by calculating the similarity between target samples and prototypes obtained from the pre-trained source model. IGDT then augments the intermediate domain samples with the target domain samples. Finally, the adaptation of the model to the target domain is further strengthened by CVCL. The overall architecture of the proposed SIDE framework is shown in Fig. \ref{fig2}.
\subsection{Problem Setup}
In a source-free domain adaptation task, the labeled source domain data with $n_s$ labeled samples is denoted as $D_s = \{{(x_s^{i},y_s^{i})}\}_{i=1}^{n_s}$, where $x_s^{i} \in \mathcal{X}_s$, ${y}_s^{i} \in \mathcal{Y}_s \subseteq \mathbb{R}^K$ and $y_s^{i}$ is the one-hot ground-truth label with $K$ dimensions. For brevity, unless otherwise stated, we neglect the superscript $i$ such as denoting $x_s^i$ as $x_s$. The unlabeled target domain data with $n_t$ samples is denoted as $D_t = \{{x_t^{i}}\}_{i=1}^{n_t}$, where $x_t^{i} \in \mathcal{X}_t$. The label space of the two domains is shared while the marginal distribution is different. The source model, denoted as $f_s(\cdot)=h_s(g_s(\cdot))$, consists of feature extractor $g_s(\cdot):\mathcal{X}_s\to \mathbb{R}^D$ and classifier $h_s(\cdot):\mathbb{R}^D\to \mathbb{R}^K$ where $D$ is the feature dimension. The source model is obtained by training the source data with the standard cross-entropy loss $L_s^{ce}= -\sum_{k=1}^K{\widetilde{y}_s^k{\rm log}p_s^k}$, where $p_s^k = \sigma_k(f_s(x_s))$ is the $k$-th element of the model's output after the softmax operation $\sigma_k(a) = \frac{{\rm exp}(a_k)}{\sum_{j=1}^K{{\rm exp}(a_j)}}$, and $\widetilde{y}_s^k$ is the $k$-th element of the converted one-hot label with label-smoothing \cite{szegedy2016rethinking}: $\widetilde{y}_s^k = (1-\tau)y_s^k + \tau/K$, where $\tau$ is the smoothness coefficient ($\tau=0.1$ in our experiment). SFDA aims to predict the labels $\{y_t^{i}\}_{i=1}^{n_t}$ in the target domain where $y_t^{i} \in  \mathcal{Y} \subseteq \mathbb{R}^K$  by learning a target model $f_t(\cdot):\mathcal{X}_t \to \mathcal{Y}_t$ based on the source model. Similar to the formula for the source model, the target model is defined as $f_t(\cdot)=h_t(g_t(\cdot))$.

\subsection{Cycle Intermediate Domain Filtering}

In order to reduce the distribution gap between the absent source data and the target data, we utilize intermediate domain samples which conform to the distribution characteristics of the target data while sharing high similarities with the source data. Classification information is hidden in the source classifier $h_s(\cdot)$ \cite{xu2020generative} and the prototype of each class can be represented by the weight vector $W_s=\{{w_k};{w_k\in \mathbb{R}^{D\times K}\}}_{k=1}^K$ of $h_s(\cdot)$ \cite{saito2018maximum}. We cyclically select the intermediate domain samples $\{\mathcal{X}_m,\mathcal{Y}_m\}={\{x_m^{i},y_m^{i}\}_{i=1}^{n_m}}$ by calculating their similarities with the prototypes, where $n_m$ is the number of intermediate samples:
\begin{equation}
\mathcal{X}_m^{k, \mathcal{C}}=
\begin{cases}
x_{m'}^{k, \mathcal{C}}, m'\in\mathop{\rm min} \limits_{t} {D_{\rm sim}(g_t(x_t),w_k)}, & \rm{if} \; \mathcal{C}=\mathcal{T}\\
\mathcal{X}_m^{k, \mathcal{C}-1},& \rm{otherwise}
\tag{1}
\end{cases}
  \label{eq1}
\end{equation}
where $D_{\rm sim}(\cdot, \cdot)$ represents the distance metric, for which we use the cosine distance by default. $\mathcal{T}$ and $\mathcal{C}$ denote the epoch value of the selected intermediate samples and the current training epoch value, respectively. $\rm min(\cdot)$ indicates that the top-$n_m$ samples most similar to the prototype in each class are selected as they have the minimum distance with $W_s$. As the training progresses, the gap between the source domain represented by the intermediate domain and target domain will gradually decrease because the distribution of the features extracted by the target model $g_{t}(\cdot)$ gradually approaches the target domain. We denote $E$ as the maximum epoch, and the target model $g_t(\cdot)$ at epoch $\mathcal{T}$ is used to filter out the intermediate samples. $\mathcal{T}$ is defined as
\begin{equation}
\begin{split}
{\mathcal{T}} = e \cdot \alpha E \quad s.t. \; 0<\mathcal{T}<E,
  \label{eq2}
\end{split}
\tag{2}
\end{equation}
where $e=\{1,2,\cdots, \lfloor \frac{1}{\alpha} \rfloor \}$ is the number of the cycles and $\alpha$ is the ratio falling in [0, 1].

Similar to the source data, we directly employ the cross-entropy loss with label smoothing as follows:
\begin{equation}
\begin{split}
\begin{aligned}
&L_{\rm int}(f_t;\mathcal{X}_m,\mathcal{Y}_m) =
&- \gamma \mathbb{E}_{(x_m,y_m)\in{X_m \times Y_m }}\sum _{k=1}^K{l_m^k {\rm log}\sigma_k(f_t(x_m^{k,\mathcal{C}}))},
  \label{eq3}
\end{aligned}
\end{split}
\tag{3}
\end{equation}
where $l_m^k = (1-\tau)y_m^k + \tau / K$ is the smoothed $k$-th element of the one-hot label and $y_m$ is the intermediate domain label corresponding to $x_m^{k, \mathcal{C}}$. $\gamma$ is the hyperparameter.

\subsection{Inter-Domain Gap Transition}

In order to make full use of the intermediate samples obtained by the progressively adapted target model, we employ mixup regularization to diffuse the effect of the intermediate samples on the entire target domain. Specifically, given two samples $x_m$, $x_t$ and these labels from the intermediate domain and target domain respectively, IDGT bridges the gap by mixing these two samples:
\begin{equation}
\begin{split}
\widehat x_u = \lambda x_m + (1-\lambda)x_t, \
\widehat q_u = \lambda q_m + (1-\lambda)\widehat q_t,
  \label{eq4}
\end{split}
\tag{4}
\end{equation}
where $\lambda \sim Beta(\beta,\beta)$ for $\beta \in (0,\infty)$, $q_m$ is the one-hot encoding of $y_m$ and $\widehat q_t$ is the soft label of $x_t$. To produce $\widehat q_t$, we build two memory banks: $B_z=[z_t^1,z_t^2,···,z_t^{n_t}]$ that stores all the features and $B_p=[p_t^1,p_t^2,...,p_t^{n_t}]$ that stores the corresponding prediction scores, where $ z_t=g(x_t)$ and $p_t = h_t(g_t(x_t))$. The historic items in the memory bank are updated corresponding to the current batch and do not require any additional computation. We then refine the target labels via nearest-neighbor soft voting by averaging the probabilities associated with the $r$-nearest neighbors ${\cal N}_r$ in the target feature space, \emph{i.e}. $q_t = \frac{1}{r}\sum_{j=1}^{r}p_{t, j}$\;, where $p_t\in{\cal N}_r$. We calculate the gap transition loss $L_{\rm gap}$ by adopting the Kullback–Leibler (KL) divergence, which can be mathematically expressed as follows:
\begin{equation}
\begin{split}
L_{\rm gap}(f_t; \widehat x_u, \widehat q_u) = \frac{1}{D}\sum _{d=1}^D{\rm KL}(\widehat{Q}_{u,(d,:)}, Q_{u,(d,:)})
=\frac{1}{D}\sum _{d=1}^D\sum _{k=1}^K \sigma_k(\widehat{q}_{u,(d,k)}{\rm log}(\frac{\widehat{q}_{u,(d,k)}}{f_t(x_{u,(d,k)})})),
\end{split}
\label{eq5}
\tag {5}
\end{equation}
where $\widehat{Q}, Q \in \mathbb{R}^{D \times K}$ are the logits before the softmax operation.

\subsection{Cross-View Consistency Learning}

To promote the feature representation learning in the target model and seek the consistent and complementary information across different views of one image, we randomly draw two augmentations $t_s$ and $t_s'$, and augment $x_t$ into two views $x^A_t=t_s(x_t)$ and $x^B_t=t_s'(x_t)$. Inspired by standard contrastive learning methods \cite{he2020momentum}, we design a momentum encoder $g_{t'}(\cdot):\mathcal{X}_t\to {\mathbb R}^D$ and a 2-layer MLP projection head $p_t(\cdot):{\mathbb R}^{D}\to {\mathbb R}^{D'}$ to support the local instance-level and global class-level consistent alignment where ${D'}$ is the output feature dimension of the 2-layer MLP projection head. The objective of this process is to minimize the following consistency loss function $L_{\rm con}$:
\begin{equation}
\begin{split}
L_{\rm con} = L_{\rm sam} + \varepsilon L_{\rm cls},
  \label{eq6}
\end{split}
\tag {6}
\end{equation}
where $ L_{\rm sam}$ is the local sample-level consistency loss and $L_{\rm cls}$ is the global class-level consistency loss, which are defined in the following. $\varepsilon$ is the trade-off parameter.

\subsubsection{Local Sample-Level Consistency}

Using a learnable non-linear projection head by contrastive loss has shown promising results for representation learning. To maximize the similarity between the two output embedding vectors of the projection heads, the feature embeddings obtained by encoders $z_t = g_t(x_t)$ and $z_{t'} = g_{t'}(x_t)$ are used to optimize the model with the following loss:
\begin{equation}
\begin{split}
L_{\rm sam}(p_t; z_{t}, z_{t'})=D_{\rm NMSE}(p_t(z_t^A), p_t(z_{t'}^B))+D_{\rm NMSE}(p_t(z_{t'}^A), p_t(z_{t}^B)),
  \label{eq7}
\end{split}
\tag {7}
\end{equation}
where $D_{\rm NMSE}(a,b) = ||\xi(a)-\xi(b)||_2^2$ is the MSE loss with the $l_2$-normalized input. As shown above, the symmetric loss is used to compute two losses by swapping two data augmentations.

 \begin{algorithm}[!t]
  \caption{The training procedure of the proposed SIDE.}
  \label{alg:Framwork}
  \begin{algorithmic}[1]
    \Require
      Target dataset $\mathcal{X}_t$, well-trained source model $f_s=h_s(g_s(\cdot))$, hyperparameters $e$, $\alpha$, $\gamma$, $\varepsilon$.
    \Ensure
      Target model $f_t = h_t(g_t(\cdot))$.
    \State Initialize $f_t$ and $g_{t'}$ with $f_s$ and $g_t$ respectively. Initialize $p_t$ randomly.
    \For{ep = 1 to $E$}
        \If {ep mod ($e \cdot \alpha)$ = 0}
            \State Sample the intermediate data $\mathcal{X}_m$ using Eq. (\ref{eq1}).
            \State Calculate $L_{\rm int}$ using Eq. (\ref{eq3}).
        \EndIf
        \State Generate pseudo-labels.
        \State Calculate $L_{\rm gap}$, $L_{\rm sam}$ and $L_{\rm cls}$ using Eq. (\ref{eq5}), Eq. (\ref{eq7}) and Eq. (\ref{eq9}), respectively.
        \State Update momentum encoder $g_{t'}$ using Eq. (\ref{eq8}).
        \State Update encoder $g_t$ and projector $p_t$ using Eqs. (\ref{eq3}), (\ref{eq5}) and (\ref{eq6}).
    \EndFor\\
    \Return Target model $f_t = h_t(g_t(\cdot))$.
  \end{algorithmic}
\end{algorithm}

In addition, the parameters $\theta_{t'}$ of the momentum model $g_{t'}$ are updated with the momentum $\omega$ and the parameters $\theta_t$ of $g_t$ at each mini-batch step instead of back-propagation:
\begin{equation}
\begin{split}
\theta_{t'} \leftarrow \omega\theta_{t'}+(1-\omega)\theta_t.
  \label{eq8}
\end{split}
\tag {8}
\end{equation}

\subsubsection{Global Class-Level Consistency}
Existing self-supervised consistency methods consider that the final predicted labels of two images should be as closer as possible if they are identical. However, these methods are limited to instance-level prediction consistency without learning compact and discriminative classification information at the global class-level. The prediction of mini-batch $ { {P_t}^A}=h_t(g_t(x_t^A)) \in \mathbb{R}^{n \times K}$ and ${P_{t'}^B}=h_t(g_{t'}({x_t^B})) \in \mathbb{R}^{n \times K}$ of views A and B are obtained by the encoder $g_t(\cdot)$ and the momentum encoder $g_{t'}(\cdot)$, followed by classifier $h_t(\cdot)$. The cross-view correlation matrix $\mathcal{ M}$ of different encoders is obtained with $\mathcal {M}= {{P_t^A}^T P_{t'}^B} \in \mathbb{R}^{K \times K}$. Inspired by \cite{yan2022multi}, we calculate the contrastive clustering loss for the predicted values of each mini-batch:
\begin{equation}
\begin{split}
\begin{aligned}
L_{\rm cls}(f_t; \mathcal{M})=\frac{{\mathcal L_1}(\xi({\mathcal{M}}),{I_n})+{\mathcal L_1}(\xi (\mathcal{M}^T),{I_n})}{2K},
  \label{eq9}
\end{aligned}
\end{split}
\tag {9}
\end{equation}
where $\mathcal L_1(\cdot)$ is the $\mathcal L_1$ loss and ${{I_n} \in {\mathbb R}^{K \times K}}$ is the identity matrix. $\xi(\cdot)$ indicates the normalization function of the matrix. Given coherence constraints on the diagonal matrix and the predicted cross-correlation matrix, SIDE can promote intra-class compactness and inter-class separability.

Algorithm \ref{alg:Framwork} demonstrates the whole training procedure of our proposed SIDE.


\begin{table}[!t]
\renewcommand\arraystretch{1}
\centering
\resizebox{0.8\linewidth}{!}{
  \begin{tabular}{lcccccccl}
    \toprule
    Methods  & A→D & A$\rightarrow$\textup  W & D$\rightarrow$\textup    A & D$\rightarrow$\textup   W & W$\rightarrow$\textup  A & W$\rightarrow$\textup D & Avg. \\
    \midrule
    Source-only  & 80.9 & 75.6 & 59.5 & 92.8 & 60.5 & 99.2  & 78.1 \\
    \midrule
    DANN\cite{ganin2015unsupervised}         & 79.7 & 82.0 & 68.2 & 96.9 & 67.4 & 99.1  & 82.2 \\
    CDAN\cite{long2018conditional}       & 92.9 & 94.1 & 71.0 & 98.6 & 69.3 & 100.0 & 87.7 \\
    SRDC\cite{tang2020unsupervised}        & 95.6 & 95.7 & 76.7 & 99.2 & 77.1 & 100.0 & 90.8 \\
    MCC\cite{jin2020minimum}         & 95.8 & 95.4 & 75.6 & 98.6 & 73.9 & 100.0 & 89.4 \\
    BDG\cite{yang2020bi}        & 93.6 & 93.6 & 73.2 & 99.0 & 72.0 & 100.0 & 88.5 \\
    BNM\cite{cui2020towards}         & 90.3 & 91.5 & 70.9 & 98.5 & 71.6 & 100.0 & 87.1 \\
    \midrule
    SHOT\cite{liang2020we}   & 94.0 & 90.1 & 74.7 & 98.4 & 74.3 & \underline{99.9}  & 88.6   \\
    BAIT\cite{yang2020unsupervised}    & 91.0 & 93.0 & 75.0 & 99.0 & 75.3 & \textbf{100.0} & 88.9   \\
    CPGA\cite{qiu2021source}    & 94.4 & 94.1 & 76.0 & 98.4 & 76.6 & 99.8  & 89.9   \\
    NRC\cite{yang2021exploiting}    & 96.0 & 90.8 & 75.3 & 99.0 & 75.0 & \textbf{100.0} & 89.4   \\
    A2Net\cite{xia2021adaptive}    & 94.5 & 94.0 & \textbf{76.7} & \textbf{99.2} & 76.1 & \textbf{100.0} &  \underline{90.1}     \\
    HCL\cite{huang2021model}     & 92.5 & \underline{94.7} & 75.9 & 98.2 & \textbf{77.7} & \textbf{100.0} & 89.8   \\
    AAA\cite{li2021divergence}   & 95.6 & 94.2 & 75.6 & 98.1 & 76.0 & 99.8 & 89.9   \\
    DIPE\cite{wang2022exploring}    & \underline{96.6} & 93.1 & 75.5 & 98.4 & \underline{77.2} & 99.6  & \underline{90.1}   \\
    \midrule
        \textbf{SIDE}& \textbf{97.1} & \textbf{97.2} & \underline{76.5} & \underline{99.1} & 76.9 & \textbf{100.0} & \textbf{91.1}\\           								
  \bottomrule
\end{tabular}
}
\caption{Classification accuracy (\%) on the \textbf{Office-31} dataset. `A→D' means A is the source domain and D is the target domain. The best and second-best results are shown in bold and with underline, respectively. Same notations are used in all tables.}
    \label{tab1}
\end{table}

\section{Experiments}
\label{sec:formatting}
\subsection{Datasets}
We evaluate our method on three benchmark datasets. \textbf{Office-31} \cite{saenko2010adapting} is composed of 4,652 images from 31 categories of everyday objects. The dataset consists of 3 subsets: Amazon (\textbf{A}), Webcam (\textbf{W}), and DSLR (\textbf{D}). \textbf{Office-Home} \cite{venkateswara2017deep} is a challenging medium-sized benchmark with 65 categories and a total of 15,500 images. The dataset consists of 4 domains: Art (\textbf{A}), Clipart (\textbf{C}), Product (\textbf{P}), and Real-world (\textbf{R}). \textbf{VisDA-C} \cite{peng2017visda} is another challenging and large-scale dataset with 12 categories in two domains. Its source domain contains 152k synthetic images generated by 3D rendering while the target domain has 55k real object images.

\begin{table*}[!t]
\renewcommand\arraystretch{0.8}
\centering
  \resizebox{0.99\linewidth}{!}{
  \begin{tabular}{lcccccccccccccl}
    \toprule
    Methods & Reference & A$\rightarrow$ \textup C & A$\rightarrow$ \textup   P & A$\rightarrow$ \textup   R & C$\rightarrow$ \textup   A &  C$\rightarrow$ \textup   P & C$\rightarrow$ \textup   R & P$\rightarrow$ \textup   A &  P$\rightarrow$ \textup   C & P$\rightarrow$ \textup   R & R$\rightarrow$ \textup   A &  R$\rightarrow$ \textup   C & R$\rightarrow$ \textup   P & Avg. \\
    \midrule
    Source-only & -  & 38.1 & 58.9 & 69.4 & 48.3 & 57.7 & 61.1 & 49.4 & 36.4 & 69.9 & 64.7 & 43.7 & 75.4 & 56.1          \\
    \midrule
    DANN\cite{ganin2015unsupervised} & PMLR'15          & 45.6 & 59.3 & 70.1 & 47.0 & 58.5 & 60.9 & 46.1 & 43.7 & 68.5 & 63.2 & 51.8 & 76.8 & 57.6          \\
    CDAN\cite{long2018conditional} & NIPS'18          & 50.7 & 70.6 & 76.0 & 57.6 & 70.0 & 70.0 & 57.4 & 50.9 & 77.3 & 70.9 & 56.7 & 81.6 & 65.8          \\
    SRDC\cite{tang2020unsupervised} & CVPR'20          & 52.3 & 76.3 & 81.0 & 69.5 & 76.2 & 78.0 & 68.7 & 53.8 & 81.7 & 76.3 & 57.1 & 85.0 & 71.3          \\
    BDG\cite{yang2020bi} & AAAI'20           & 51.5 & 73.4 & 78.7 & 65.3 & 71.5 & 73.7 & 65.1 & 49.7 & 81.1 & 74.6 & 55.1 & 84.8 & 68.7          \\
    BNM\cite{cui2020towards} & CVPR'20           & 52.3 & 73.9 & 80.0 & 63.3 & 72.9 & 74.9 & 61.7 & 49.5 & 79.7 & 70.5 & 53.6 & 82.2 & 67.9          \\
    \midrule
    SHOT\cite{liang2020we} & PMLR'20          & 57.1 & 78.1 & 81.5 & 68.0 & 78.2 & 78.1 & 67.4 & 54.9 & 82.2 & 73.3 & 58.8 & 84.3 & 71.8          \\
    BAIT\cite{yang2020unsupervised} & arXiv'20          & 57.4 & 77.5 & \underline{82.4} & 68.0 & 77.2 & 75.1 & 67.1 & 55.5 & 81.9 & 73.9 & 59.5 & 84.2 & 71.6          \\
    A2Net\cite{xia2021adaptive} & ICCV'21          &\underline{58.4} & 79.0 & \underline{82.4} & 67.5 & 79.3 & 78.9 & \underline{68.0} & 56.2 & 82.9 & 74.1 & 60.5 & 85.0 & \underline{72.8}     \\
    CPGA\cite{qiu2021source} & IJCAI'21          & \textbf{59.3} & 78.1 & 79.8 & 65.4 & 75.5 & 76.4 & 65.7 & \textbf{58.0} & 81.0 & 72.0 & \textbf{64.4} & 83.3 & 71.6          \\
    NRC\cite{yang2021exploiting} & NIPS'21           & 57.7 & \textbf{80.3} & 82.0 & \underline{68.1} & \textbf{79.8} & 78.6 & 65.3 & \underline{56.4} & 83.0 & 71.0 & 58.6 & \textbf{85.6} & 72.2          \\
    AAA\cite{li2021divergence} & TPAMI'21          & 56.7 & 78.3 & 82.1 & 66.4 & 78.5 & \underline{79.4} & 67.6 & 53.5 & 81.6 & \textbf{74.5} & 58.4 & 84.1 & 71.8 \\
    DIPE\cite{wang2022exploring} & CVPR'22          & 56.5 & 79.2 & 80.7 & \textbf{70.1} & \textbf{79.8} & 78.8 & 67.9 & 55.1 & \underline{83.5} & 74.1 & 59.3 & 84.8 & 72.5          \\
    \textbf{SIDE} & Ours & 56.9 & \underline{79.7} & \textbf{82.5} & \textbf{70.1} & \underline{79.4} & \textbf{80.0} & \textbf{68.5} & 55.2 & \textbf{84.0} & \underline{74.3} & \underline{60.6} & \underline{85.2} & \textbf{73.0}\\							
  \bottomrule
\end{tabular}
}
\vspace{-2mm}
\caption{Classification accuracy (\%) on the \textbf{Office-Home} dataset.}
  \label{tab2}
\end{table*}

\begin{table*}[!t]
\renewcommand\arraystretch{0.8}
\centering 
\resizebox{\textwidth}{!}{
  \resizebox{0.99\linewidth}{!}{
  \begin{tabular}{lcccccccccccccl}
    \toprule
    Methods & Reference & plane & bike & bus & cat & horse & kniffe & mcycle & person & plant & sktbrd & train & truck & Avg. \\
    \midrule
    Source-only & - & 67.9 & 11.1 & 57.7 & 70.9 & 63.3 & 8.9  & 79.1 & 21.6 & 68.1 & 16.8 & 84.6 & 9.4  & 46.6 \\
    \midrule
    DANN\cite{ganin2015unsupervised} & PMLR`15        & 81.9 & 77.7 & 82.8 & 44.3 & 81.2 & 29.5 & 65.1 & 28.6 & 51.9 & 54.6 & 82.8 & 7.8  & 57.4 \\
    CDAN\cite{long2018conditional} & NIPS'18        & 85.2 & 66.9 & 83.0 & 50.8 & 84.2 & 74.9 & 88.1 & 74.5 & 83.4 & 76.0 & 81.9 & 38.0 & 73.9 \\
    SWD\cite{lee2019sliced} & CVPR'19         & 90.8 & 82.5 & 81.7 & 70.5 & 91.7 & 69.5 & 86.3 & 77.5 & 87.5 & 63.6 & 85.6 & 29.2 & 76.4 \\
    MCC\cite{jin2020minimum} & ECCV'20         & 88.7 & 80.3 & 80.5 & 71.5 & 90.1 & 93.2 & 85.0 & 71.6 & 89.4 & 73.8 & 85.0 & 36.9 & 78.8 \\
    CoSCA\cite{dai2020contrastively} & ACCV'20       & 95.7 & 87.4 & 85.7 & 73.5 & 95.3 & 72.8 & 91.5 & 84.8 & 94.6 & 87.9 & 87.9 & 36.9 & 82.9 \\
    PAL\cite{hu2020panda} & arXiv'20         & 90.9 & 50.5 & 72.3 & 82.7 & 88.3 & 88.3 & 90.3 & 79.8 & 89.7 & 79.2 & 88.1 & 39.4 & 78.3 \\
    \midrule
    SHOT\cite{liang2020we} & PMLR'20        & 94.3 & \textbf{88.5} & 80.1 & 57.3 & 93.1 & 94.9 & 80.7 & 80.3 & 91.5 & 89.1 & 86.3 & 58.2 & 82.9 \\
    BAIT\cite{yang2020unsupervised} & arXiv'20        & 93.7 & 83.2 & \underline{84.5} & 65.0 & 92.9 & \textbf{95.4} & \underline{88.1} & 80.8 & 90.0 & 89.0 & 84.0 & 45.3 & 82.7 \\
    A2Net\cite{xia2021adaptive} & ICCV'21        & 94.0  & \underline{87.8} & \textbf{85.6} & 66.8 & 93.7 & \underline{95.1} & 85.8 & \underline{81.2} & 91.6 & 88.2 & 86.5 & 56.0 & \underline{84.3} \\
    HCL\cite{huang2021model} & NIPS'21         & 93.3 & 85.4 & 80.7 & \underline{68.5} & 91.0 & 88.1 & 86.0 & 78.6 & 86.6 & 88.8 & 80.0 & \textbf{74.7} & 83.5 \\
    AAA\cite{li2021divergence} & TPAMI'21          & 94.4 & 85.9 & 74.9 & 60.2 & \textbf{96.0} & 93.5 & 87.8 & 80.8 & 90.2 & \underline{92.0} & \underline{86.6} & \underline{68.3} & 84.2 \\
    DIPE\cite{wang2022exploring} & CVPR'22        & \underline{95.2} & 87.6 & 78.8 & 55.9 & \underline{93.9} & 95.0 & 84.1 & \textbf{81.7 }& \underline{92.1} & 88.9 & 85.4 & 58.0 & 83.1 \\
    \textbf{SIDE} & Ours & \textbf{95.9} & 84.3 & 83.7 & \textbf{68.7} & \underline{93.9} & 92.2 & \textbf{91.3} & 79.3 & \textbf{93.3} & \textbf{93.2} & \textbf{88.2} & 56.8 & \textbf{85.1}	\\				
  \bottomrule
\end{tabular}
}}
\vspace{-2mm}
\caption{Classification accuracy (\%) on the \textbf{VisDA-C} dataset.}
    \label{tab3}
\end{table*}
\subsection{Compared Methods}
We compare our method with different kinds of state-of-the-art methods, including 1) source-only: ResNet \cite{he2016deep}; 2) unsupervised domain adaptation with source data: DANN\cite{ganin2015unsupervised}, CDAN\cite{long2018conditional}, SRDC\cite{tang2020unsupervised}, MCC\cite{jin2020minimum}, BDG\cite{yang2020bi}, BNM \cite{cui2020towards}, SWD\cite{lee2019sliced}, CoSCA\cite{dai2020contrastively}, and PAL\cite{hu2020panda}; 3) source-free unsupervised domain adaptation: SHOT\cite{liang2020we}, BAIT\cite{yang2020unsupervised}, CPGA\cite{qiu2021source}, NRC\cite{yang2021exploiting}, BDT\cite{kundu2022balancing},    A2Net\cite{xia2021adaptive}, HCL\cite{huang2021model}, AAA\cite{li2021divergence}, and DIPE\cite{wang2022exploring}.

\subsection{Implementation Details}
All the experiments are conducted using the PyTorch library. For fair comparison, we also use ResNet-50 as the backbone for the Office datasets and ResNet-101 \cite{he2016deep} for VisDA-C as the feature encoder. The projection head is a 2-layer MLP and the classifier follows SHOT \cite{liang2020we} to add a bottleneck consisting of a fully-connected layer, followed by a batch normalization layer. We train the source model using the SGD optimizer with the learning rate of 1e-3 for the backbone and 1e-2 for the classifier. In the adaptation stage, we set the learning rate to 1e-3 for the backbone and 1e-2 for projection. We set the training epoch to 100, 40 and 2 for Office, Office-Home and VisDA, respectively. For hyperparameters, we set $\alpha$ and $e$ in Eq. (\ref{eq2}) as 0.3 and 2, respectively. The $\gamma$ in Eq. (\ref{eq3}) and $\epsilon$ in Eq. (\ref{eq6}) are set to 0.1 and 0.01. Following \cite{ding2022proxymix}, we adopt $n_m$ = 5, 5, 50 in Eq. (\ref{eq1}) for Office-31, Office-Home and VisDA-C, respectively. All the results are averaged over three runs with seeds $\in$ {0, 1, 2}.



\begin{figure}[htbp]
\begin{minipage}[t]{0.48\textwidth}
\centering
\includegraphics[width=\textwidth]{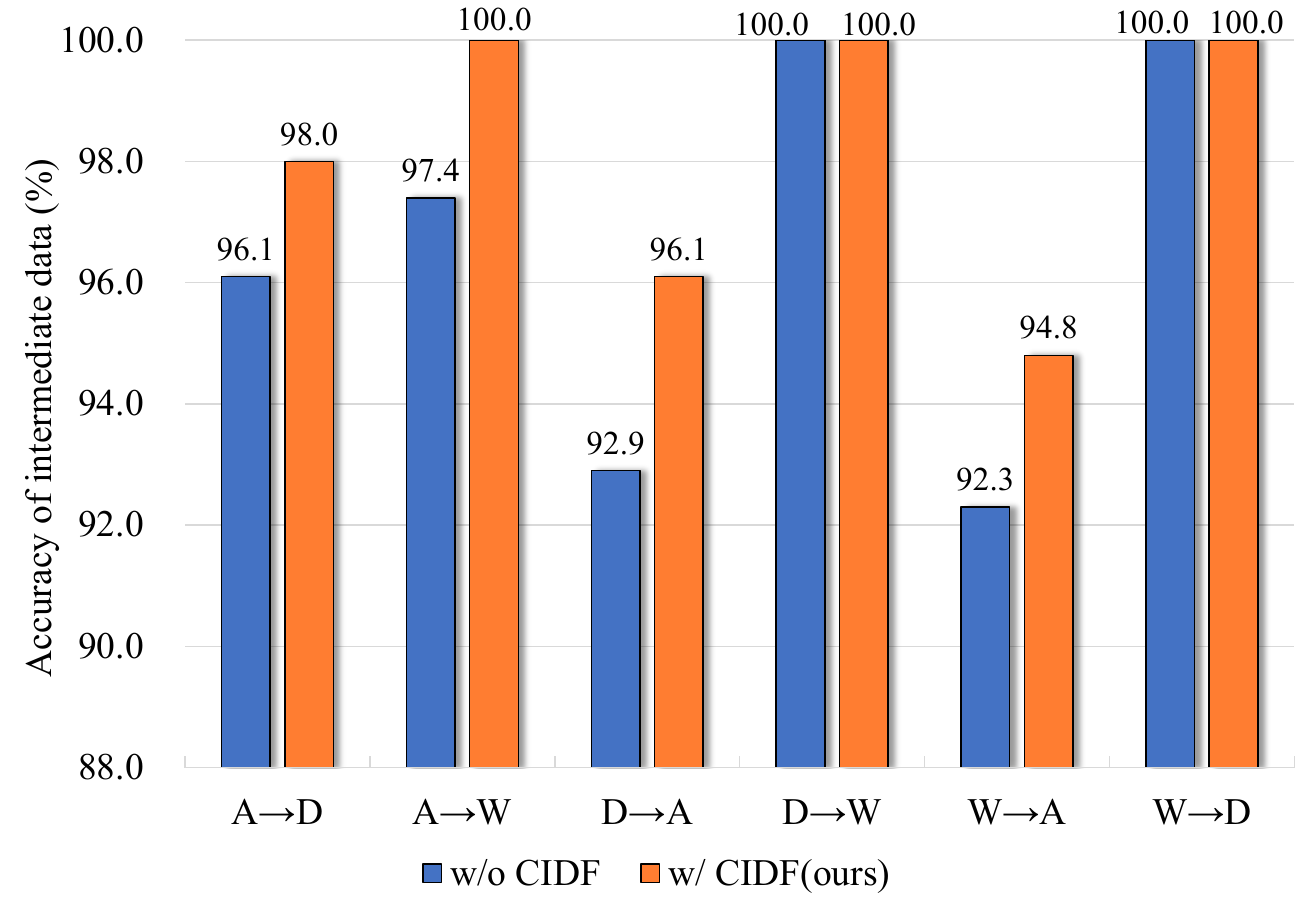}
\end{minipage}
\begin{minipage}[t]{0.48\textwidth}
\centering
\includegraphics[width=\textwidth]{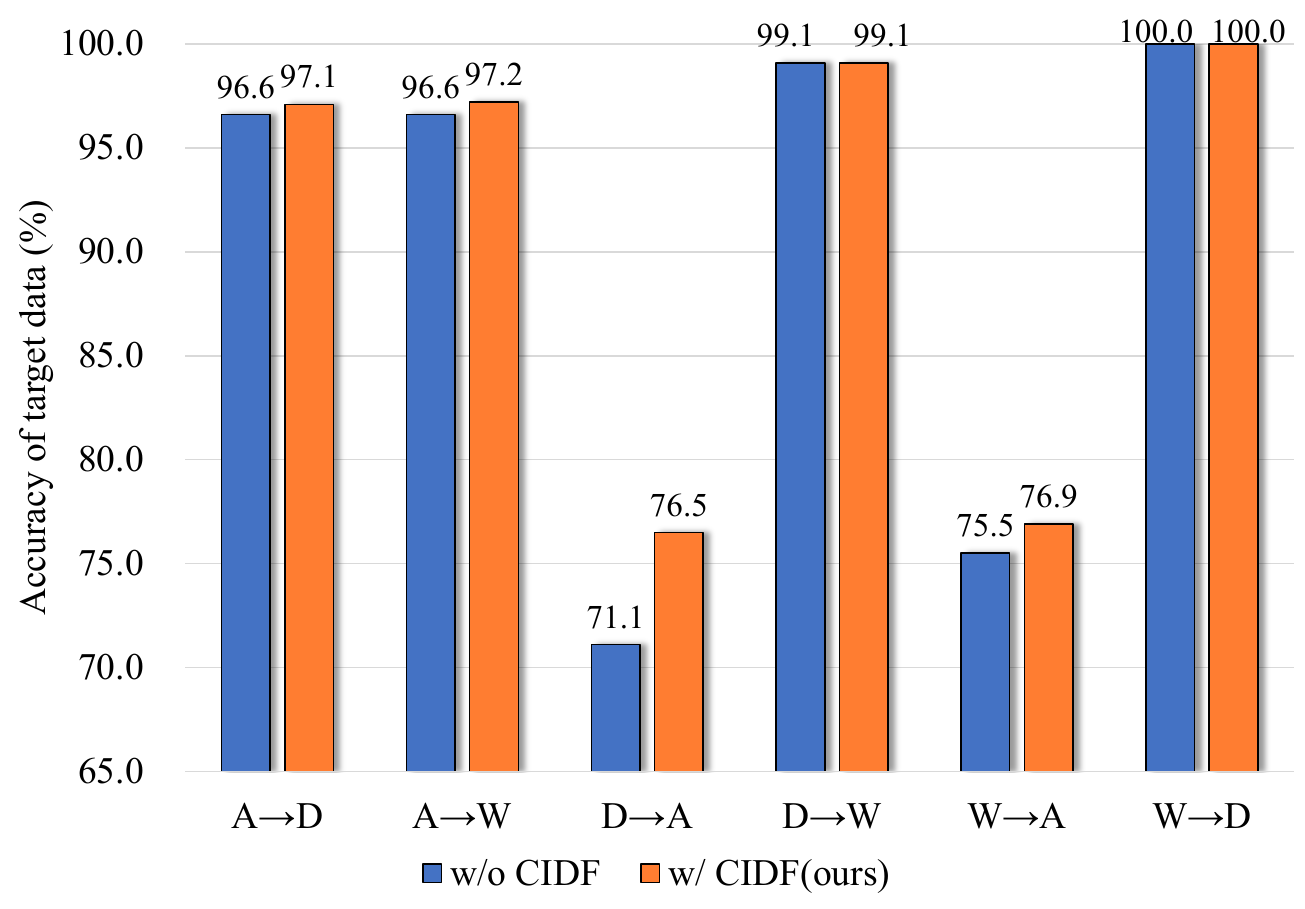}
\end{minipage}
\caption{Average accuracy of intermediate data(left) and Average accuracy on intermediate and target data. The adaptability of the target model is closely correlated with the screening ability of the intermediate data in the CIDF module(right).}
\label{fig3}
\end{figure}

\subsection{Results}
\textbf{Office-31}. On this dataset, the transfer of task D $\rightarrow$ W and W $\rightarrow$ D are relatively simple due to the small gap between D and W, while there is still much room for improvement in other tasks. As shown in Table \ref{tab1}, the proposed SIDE outperforms most state-of-the-art methods and the improvement on A $\rightarrow$ D and A $\rightarrow$ W is especially obvious. Compared to SRDC, the current best UDA method on Office-31, our proposed method gains the improvement of 0.3\%. Compared with the best result of the previous SFDA method, SIDE improves the accuracy by 1.0\%.

\textbf{Office-Home}. As we can see from Table \ref{tab2}, our method achieves the best performance in average accuracy on this dataset, compared with the other SFDA methods. The best or second-best results are achieved by our method in terms of 10 out of 12 transfer tasks. The results are consistent with our expectations that our method can also perform well on relatively large dataset.

\textbf{VisDA-C}. From Table \ref{tab3}, we can observe that SIDE achieves the best accuracy or second-best accuracy in terms of more than half of the categories and obtains comparable results in the rest categories on this large-scale dataset. The experimental results above clearly demonstrate the effectiveness of our method in addressing the multi-scale and domain-gap challenges.

In summary, our method achieves the above promising results mainly due to the introduction of self-supervised learning, which is beneficial for learning the feature distribution of the target domain. In the meantime, both CIDF and IDGT modules can effectively reduce the implicit domain gap. Different from the methods based on pseudo source domain generation, SIDE avoids changing the intrinsic structure of the target domain when aligning the target domain with the pseudo-source domain. 

\begin{minipage}[c]{0.5\textwidth}
\centering
    \scalebox{0.7}{ 
    \begin{tabular}{cccc|cc}
      \toprule
        Backbone & CIDF & IDGT & CVCL & \textrm{Office-31}  & \textrm{Office-Home}\\
        \midrule
        \checkmark &   &  &   & 78.1  &  56.1     \\
        \checkmark & \checkmark &  &   & 84.8 & 64.1 \\
        \checkmark &   &  &   \checkmark  &  87.1 &  69.2 \\
        \checkmark & \checkmark &  & \checkmark & 87.6 & 69.4 \\
        \checkmark & \checkmark & \checkmark &  & 90.8 & 71.4  \\
        \checkmark & \checkmark & \checkmark & \checkmark & \textbf{91.1} & \textbf{73.0} \\
        \bottomrule
\end{tabular}
}
\captionof{table}{Ablation study on the loss functions.}
\label{tab5}
\end{minipage}
\begin{minipage}[c]{0.5\textwidth}
\scalebox{0.8}{ 
    \begin{tabular}{*{10}{c}}
      \toprule
        \multicolumn{5}{c}{$\alpha$} & \multicolumn{4}{c}{$e$} \\
        \cmidrule(lr){1-5} \cmidrule(lr){6-10}
          & 0.1     & 0.2     & 0.3     & 0.4    & 0.5    & 1        & 2        & 3         & 4         \\
        \midrule
         & 68.7    & 69.5    & \textbf{70.1}    & 69.7   & 69.9   & 68.9       & \textbf{70.1}       & 69.1      & 69.4     \\
        \bottomrule
\end{tabular}
}
\captionof{table}{Influence of the trade-off parameter $\alpha$ and $e$ in C → A accuracy (\%) on the Office-Home dataset.}
    \label{tab4}

\end{minipage}

\begin{figure}[htbp]
\begin{minipage}[t]{0.5\textwidth}
\centering
\includegraphics[width=0.7\textwidth]{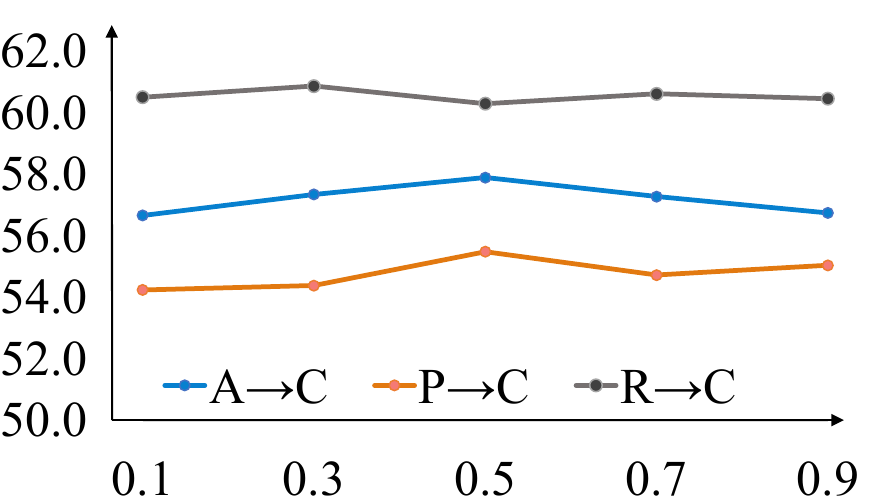}
\end{minipage}
\begin{minipage}[t]{0.5\textwidth}
\centering
\includegraphics[width=0.7\textwidth]{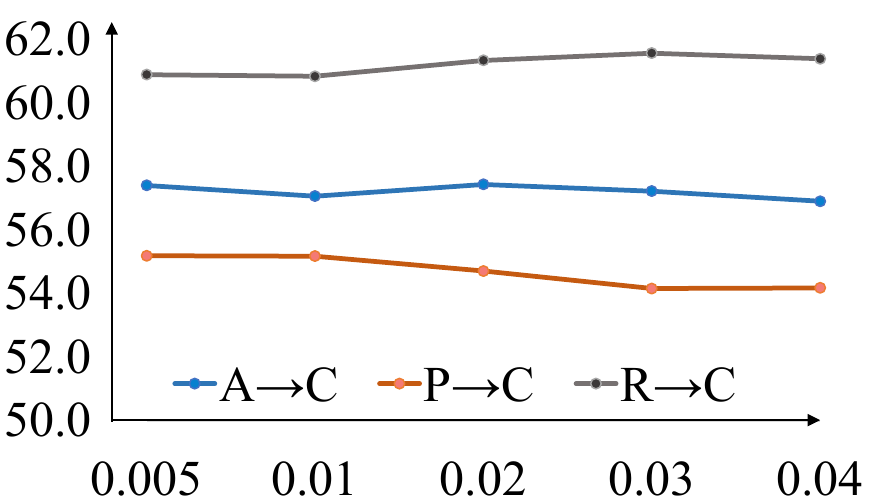}
\end{minipage}
\caption{Results of parameter $\gamma$(left) and $\varepsilon$(right) sensitivity.}
\label{fig4}
\end{figure}
\subsection{Ablation Study}

\textbf{Intermediate Samples Selection}. To verify the effectiveness of the cyclically selected intermediate samples, we compare non-cyclic filtering (\emph{i.e.} the intermediate samples are only selected once before the model training) with cyclic filtering during the training process. As Fig. \ref{fig3} shows, the experiment shows that with cyclic filtering, the accuracy of the selected intermediate domain samples is improved from 92.9\% to 96.1\% in D $\rightarrow$ A, and the corresponding test in the target domain can also obtain accuracy improvement of 5.4\%. Similarly, except for the relatively simple domain adaptation tasks W $\rightarrow$ D and D $\rightarrow$ W, the final adaptation outcome in the other tasks is corresponding to the outcome on the intermediate data. 

\textbf{Hyperparameters}. In this subsection, we evaluate the sensitivity of hyperparameters with the tasks A $\rightarrow$ C, P $\rightarrow$ C, R $\rightarrow$ C on Office-Home. Table  \ref{tab4} shows the sensitivity of performance with $e\in$ \{0.1, 0.3, 0.5, 0.7, 0.9\} and $\alpha\in$ \{0.005, 0.01, 0.02, 0.03, 0.04\} in Eq. (\ref{eq2}). The results show that our method is insensitive to both parameters. Because the proposed CIDF, IGDT and CVCL alternatively promote each other by providing positive feedback. The more optimized the model can be during the training process, the more reliable the selection of intermediate samples becomes. Furthermore, we evaluate  $\varepsilon$ and $\gamma$ on Office-Home, and as shown in Fig. \ref{fig4}, our method is not sensitive to these hyperparameters either.



\begin{figure}[htbp]
\begin{minipage}[t]{0.05\textwidth}
\centering
\includegraphics[width=\textwidth]{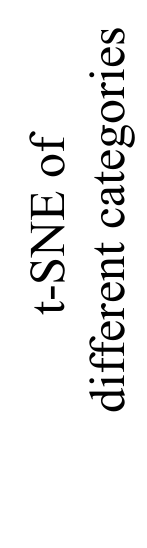}
\end{minipage}
\begin{minipage}[t]{0.30\textwidth}
\centering
\includegraphics[width=\textwidth]{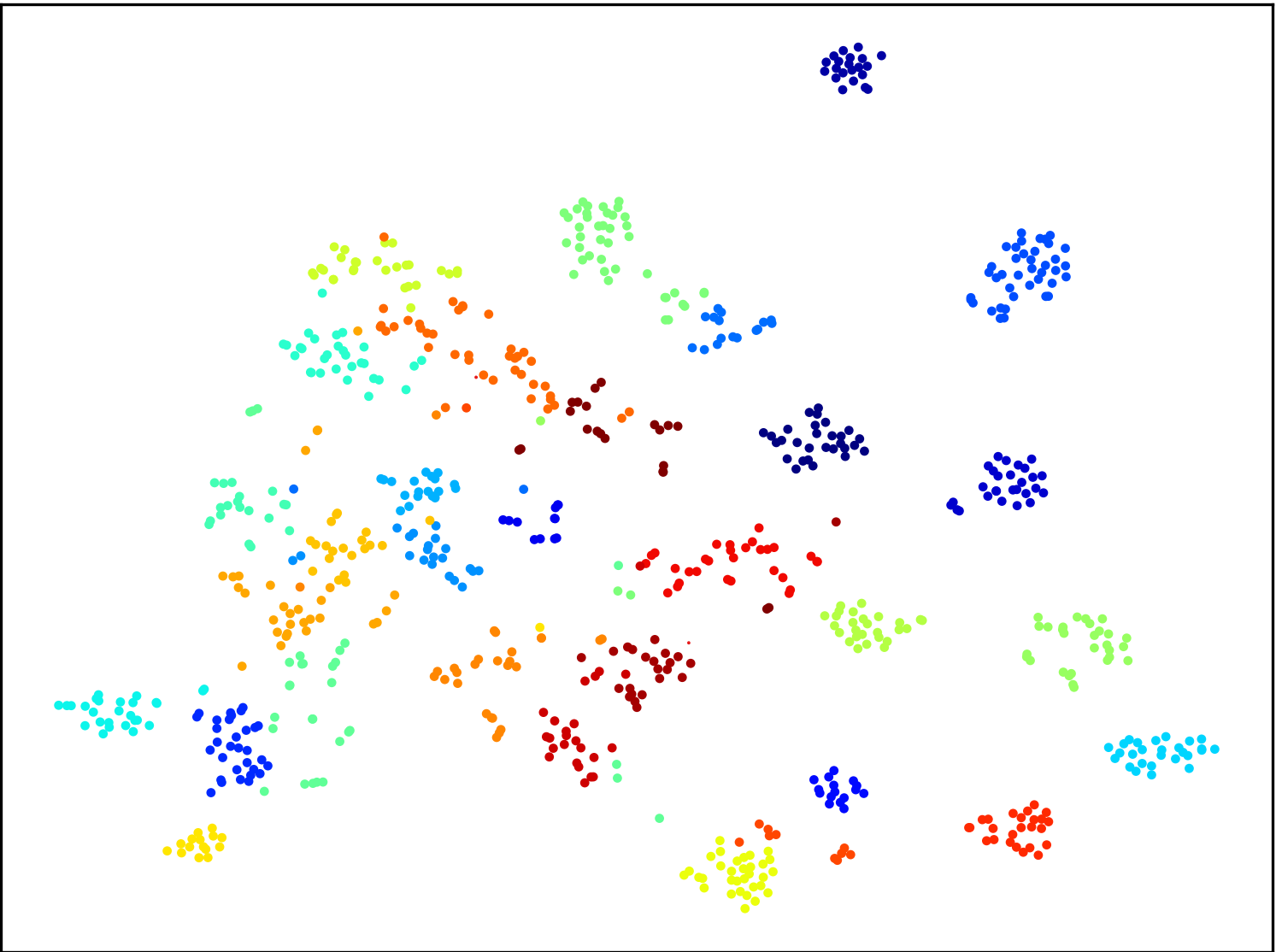}
\text{(a) Source-only}
\end{minipage}
\begin{minipage}[t]{0.30\textwidth}
\centering
\includegraphics[width=\textwidth]{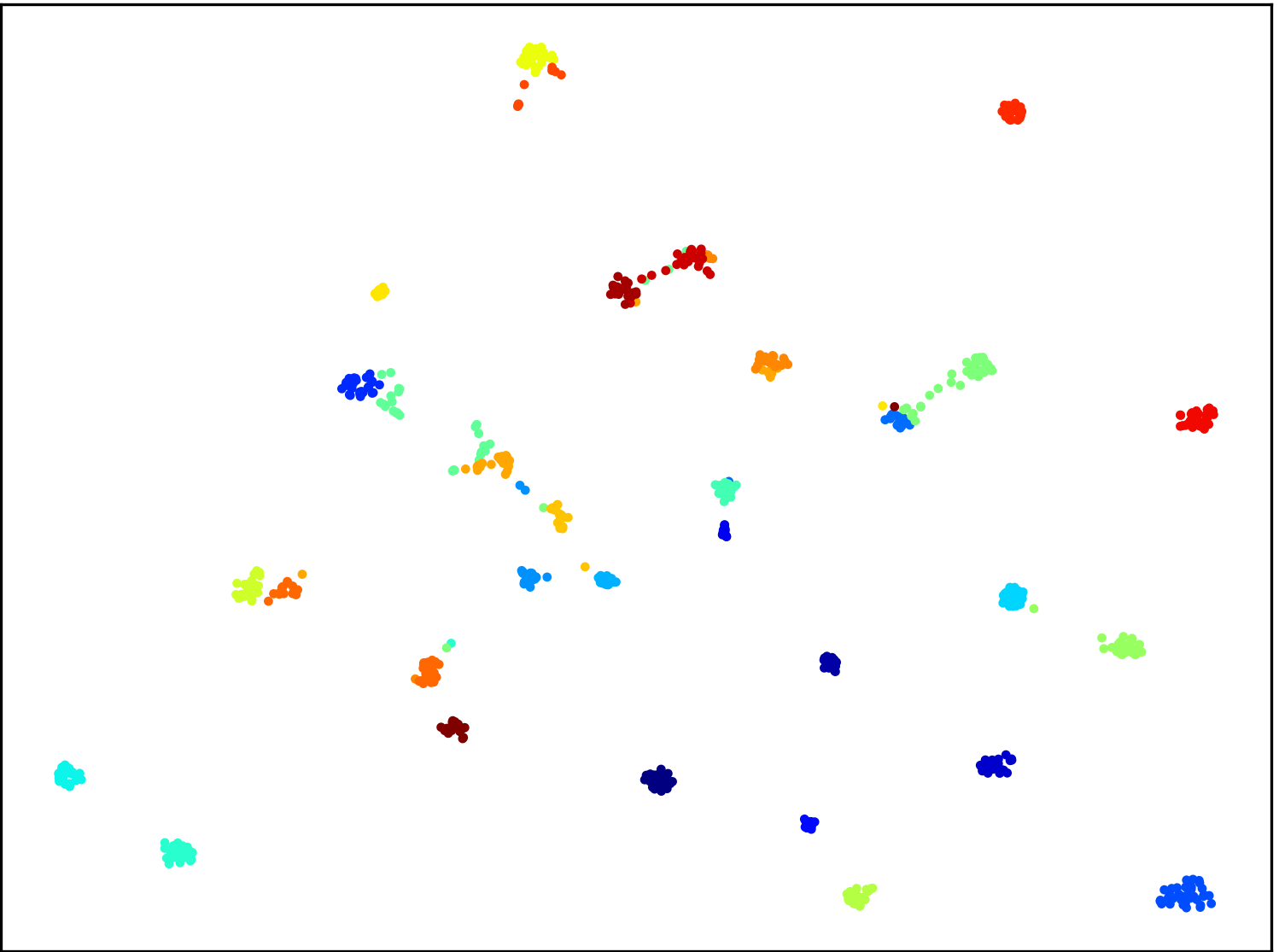}
\text{(b) SHOT}
\end{minipage}
\begin{minipage}[t]{0.30\textwidth}
\centering
\includegraphics[width=\textwidth]{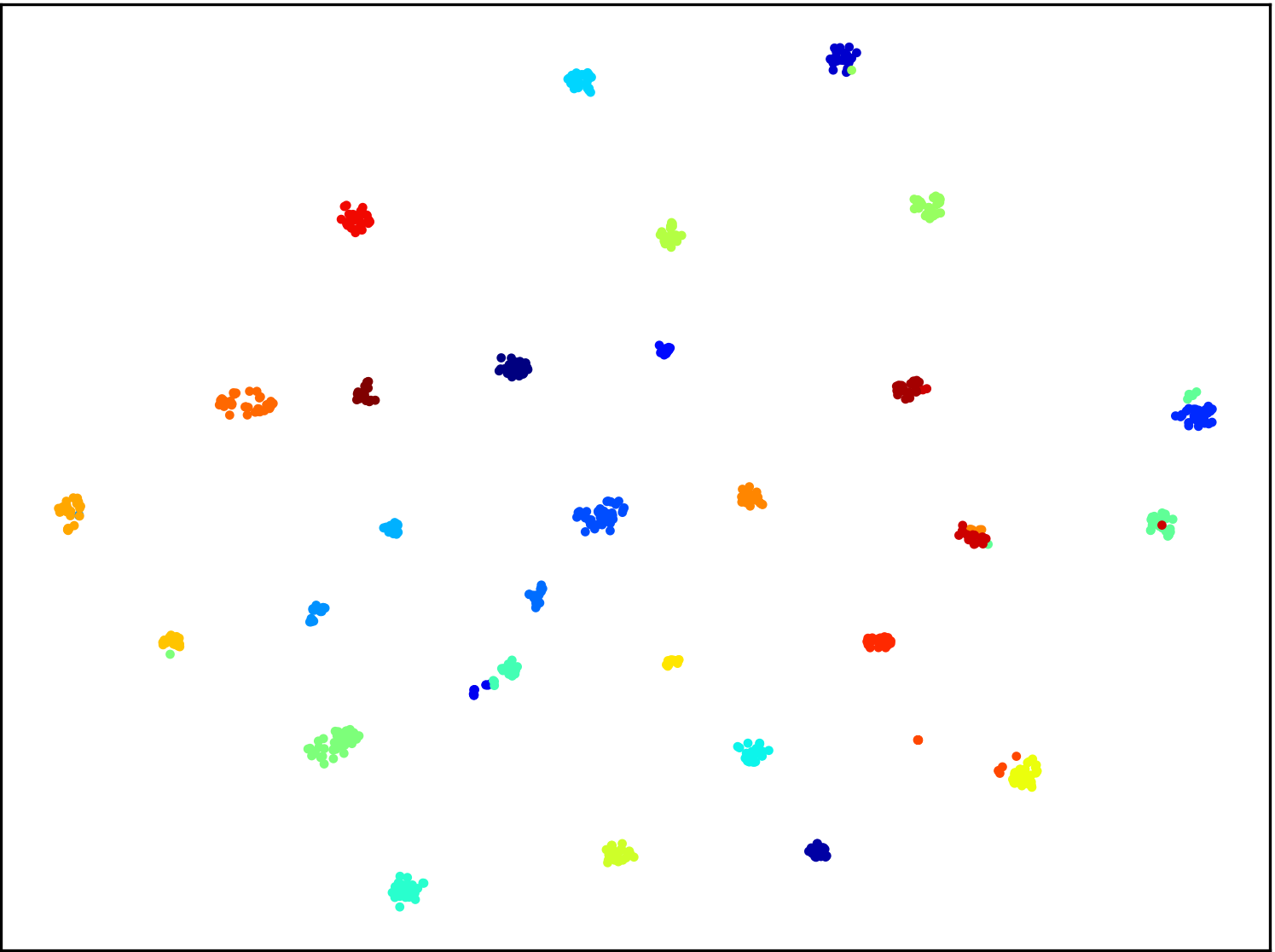}
\text{(b) SIDE}
\end{minipage}

\begin{minipage}[t]{0.05\textwidth}
\centering
\includegraphics[width=\textwidth]{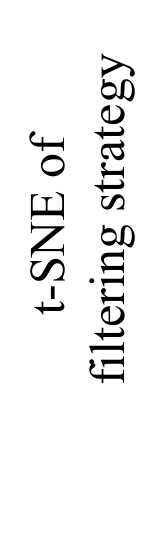}
\end{minipage}
\begin{minipage}[t]{0.30\textwidth}
\centering
\includegraphics[width=\textwidth]{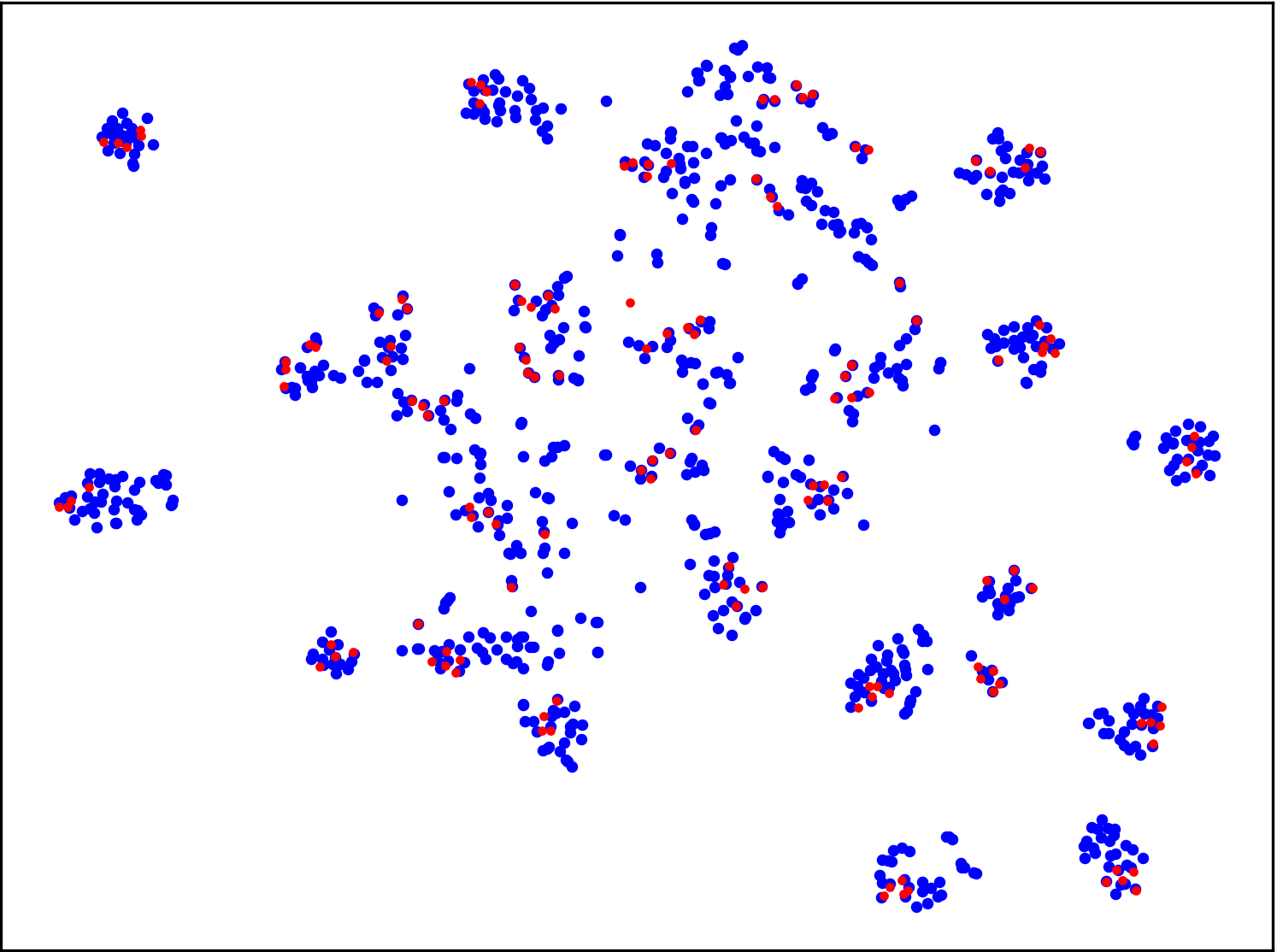}
\text{(b) Source-only}
\end{minipage}
\begin{minipage}[t]{0.30\textwidth}
\centering
\includegraphics[width=\textwidth]{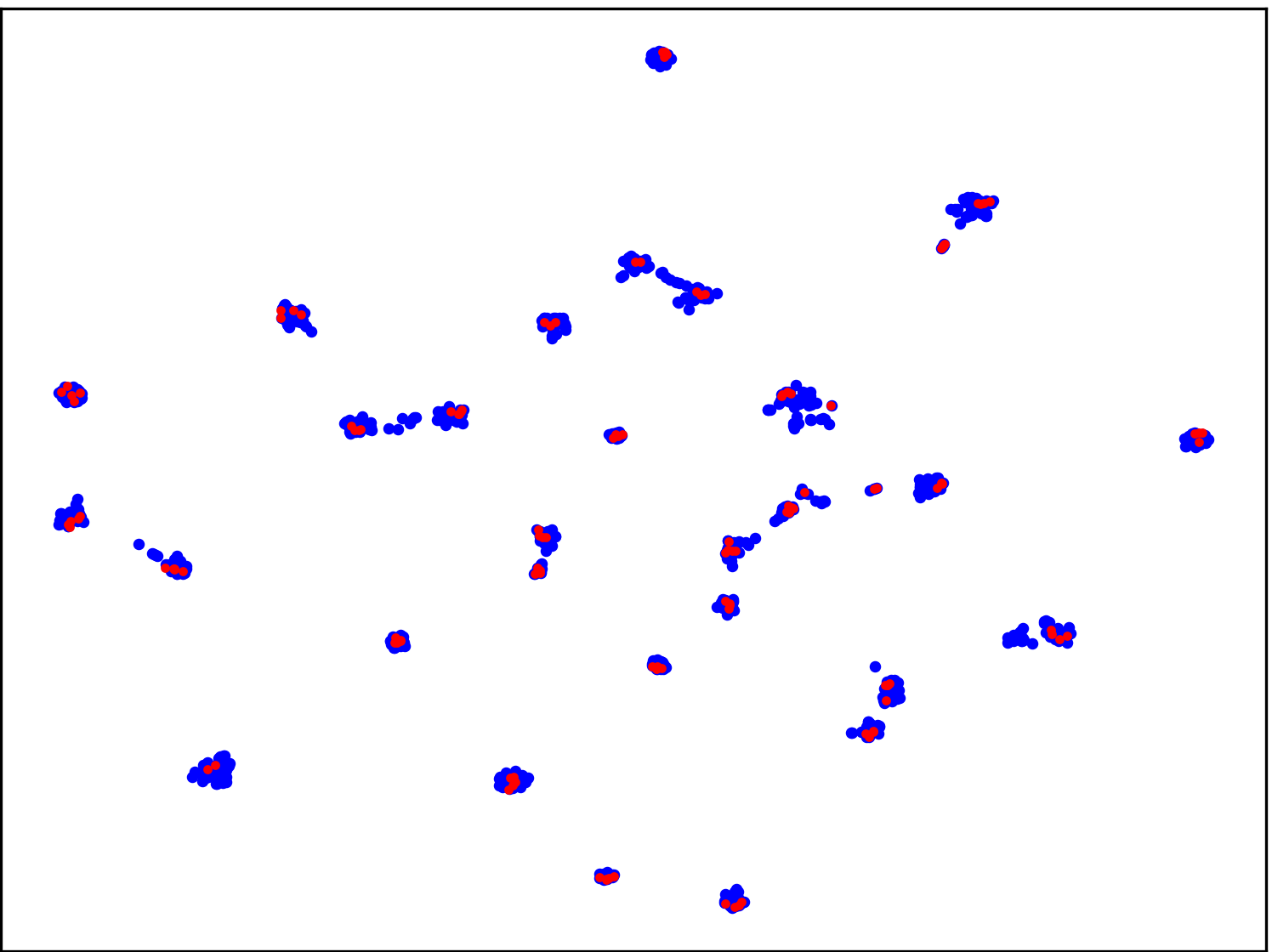}
\text{(b) SHOT}
\end{minipage}
\begin{minipage}[t]{0.30\textwidth}
\centering
\includegraphics[width=\textwidth]{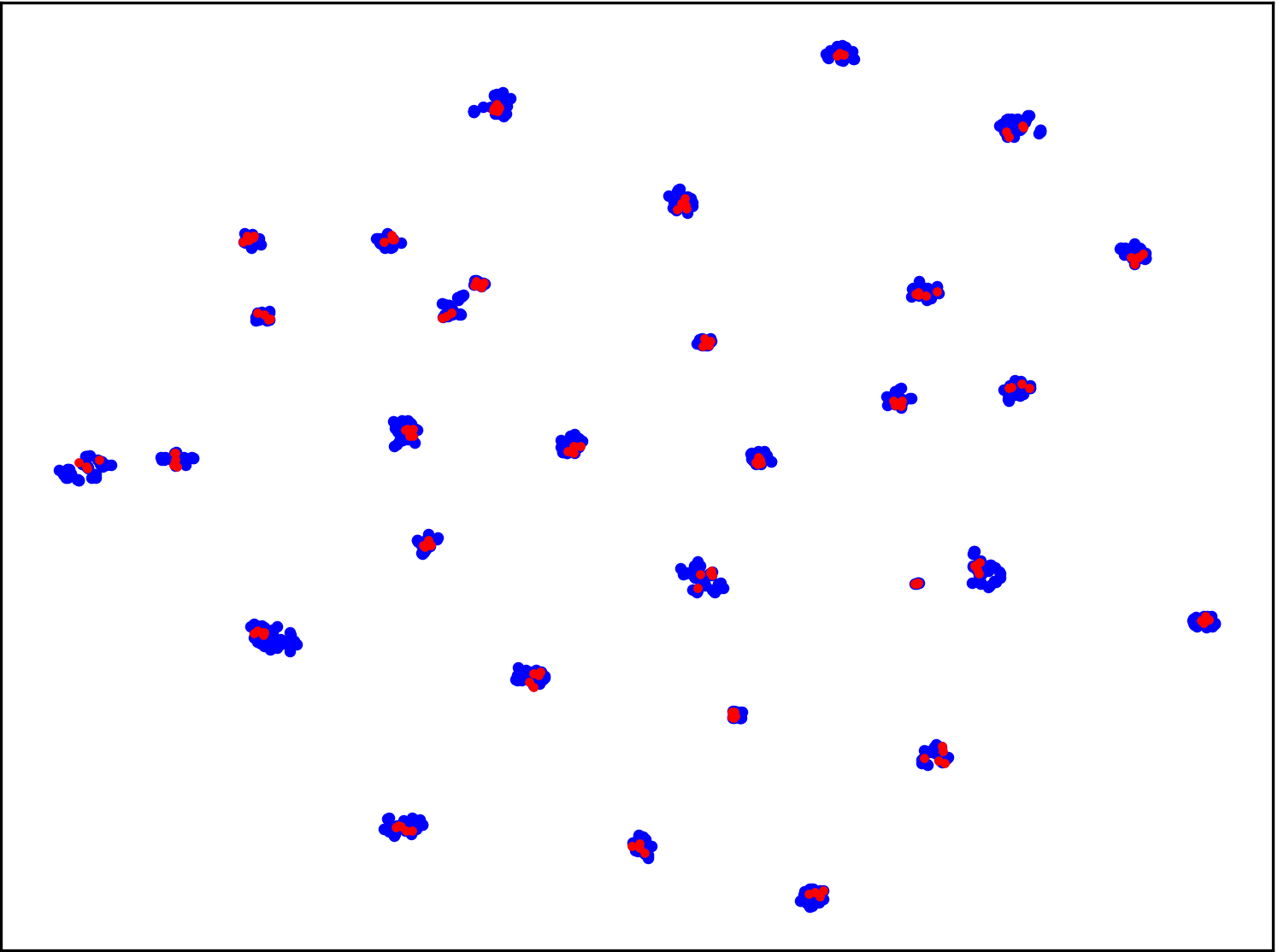}
\text{(b) SIDE}
\end{minipage}
\caption{The t-SNE visualizations of different methods on the target domain (Webcam) of the Office-31 task A → W. The first row shows the visualization of category features in the target domain where the different colors denote different categories. The second row shows the visualization of the filtered intermediate samples (in red) and other target samples (in blue). Best viewed in color.}
\label{fig5}
\end{figure}

\textbf{Component Analysis}. To investigate the effectiveness of the proposed modules in the adaptation phase, the quantitative results of the model with different components are shown in Table \ref{tab5}. 1) When only using CIDF, the performance increases by 6.7\% and 8.0\% on Office-31 and Office-Home, respectively, compared to the source-only model. The partial samples in the target domain serve as the intermediate domain to implicitly align the distributions. 2) Introducing CVCL alone encourages the model to learn a deeper internal structure of the target domain, increasing the model performance significantly. It is worth noting that IDGT cannot be used alone as it depends on the intermediate samples from CIDF. 3) Using CIDF and CVCL at the same time can improve the adaptability of the model but not achieve the optimal performance. Because the intermediate sample does not play its role at this time and the two modules lack a bridge to promote each other. 4)  Combining CIDF and IDGT, the model can better be adapted to the target domain as the intermediate data is used as a pedal to eliminate the domain gap. This combination ignores the learning of the internal distribution information of the target domain in the absence of CVCL. 5) The best results can only be achieved when combining all three modules as they reinforce each other to improve the adaptability of the target model.

\textbf{t-SNE Visualization}. We show the t-SNE feature visualization on the task A → W of target features on Office-31. In Fig. \ref{fig5} (a), the feature distribution is rather chaotic when using only the source domain model. After the adaptation of the SIDE method, as shown in Fig. \ref{fig5} (c), the features of the same category are closely related. Its classification boundary is more discriminative and the distribution is more compact compared with SHOT, as shown in Fig. \ref{fig5} (b). This is because the category information of the target domain has been deeply examined by the CVCL module through self-supervised learning, without changing the intrinsic structure of the target domain by the alignment of the two domains.

The distribution of the intermediate samples in the target domain is also visualized. As shown in Fig. \ref{fig5} (f), the intermediate domain samples obtained by CIDF gather in the center of a certain category feature. Compared with the method without any adaptation strategy in Fig. \ref{fig5} (d), the distance between intermediate samples in the same class selected by SIDE is closer. The promotion of CIDF is benefited by the CVCL module mining the semantic information of the target domain and the IDGT module strengthening the influence of intermediate samples through data augmentation. The method using other adaptation strategies in Fig. \ref{fig5} (e) can also enhance intra-class compactness to a certain extent, but there is still inter-class adhesion.

\section{Conclusion}
In this paper, we propose a novel method named SIDE to tackle the task of source-free domain adaptation via cyclically exploiting the intermediate samples. With the help of intermediate samples, SIDE effectively bridges the gap between the unseen source and target domains. The instance- and class-level consistency learning across different views gradually improves the classification capability of the target model by learning more discriminative features. Extensive experiments on three datasets show that our proposed SIDE outperforms other state-of-the-art methods.

\bibliographystyle{unsrt}  
\bibliography{references}

\end{document}